\documentclass[conference]{IEEEtran}
\IEEEoverridecommandlockouts
\usepackage{cite}
\usepackage{amsmath,amssymb,amsfonts}
\usepackage{graphicx}
\usepackage{textcomp}
\usepackage{xcolor}
\usepackage{booktabs}

\def\BibTeX{{\rm B\kern-.05em{\sc i\kern-.025em b}\kern-.08em
    T\kern-.1667em\lower.7ex\hbox{E}\kern-.125emX}}

\usepackage{subcaption}
\usepackage{booktabs}
\usepackage{gensymb} % degree symbol
\usepackage{algorithm}
\usepackage{algpseudocode} % for algorithmic environment
\usepackage{amssymb}
\usepackage{optidef}
\usepackage{adjustbox}

\algtext*{EndWhile}% Remove "end while" text
\algtext*{EndIf}% Remove "end if" text
\algtext*{EndFor}% Remove "end for" text
\algtext*{EndProcedure}% Remove "end for" text

\usepackage{microtype}

\usepackage{comment}
\usepackage{ifthen}
\newcommand{\showcomments}{yes}

\newcommand\ming[1]{
    \ifthenelse{\equal{\showcomments}{no}}{{\color{red} [Ming: #1]}}{\ignorespaces}
}
\newcommand\m[1]{
    \ifthenelse{\equal{\showcomments}{no}}{{\color{red} [Ming: #1]}}{\ignorespaces}
}
\newcommand\yt[1]{
    \ifthenelse{\equal{\showcomments}{yes}}{{\color{black} #1}}{\ignorespaces}
}

\newcommand\me[1]{
    \ifthenelse{\equal{\showcomments}{no}}{{\color{blue} #1}}{\ignorespaces}
}

\begin{document}

\title{LiDAR-based Outdoor Crowd Management for Smart Campus on the Edge\\
}

\author{}

\author{\IEEEauthorblockN{1\textsuperscript{st} Yitao Chen}
\IEEEauthorblockA{
\textit{Arizona State University}\\
ychen404@asu.edu
}
\and
\IEEEauthorblockN{2\textsuperscript{nd} Krishna Gundu}
\IEEEauthorblockA{
\textit{Arizona State University}\\
kgundu1@asu.edu}
\and
\IEEEauthorblockN{3\textsuperscript{rd} Zohair Zaidi}
\IEEEauthorblockA{
\textit{Arizona State University}\\
zohair.zaidi@asu.edu}
\and
\IEEEauthorblockN{4\textsuperscript{th} Ming Zhao}
\IEEEauthorblockA{
\textit{Arizona State University}\\
mingzhao@asu.edu}
}

% \author{\IEEEauthorblockN{1\textsuperscript{st} Yitao Chen}
% \textit{Arizona State University}\\}
% \and
% \and
% \and
% \and
% \and

\maketitle

% Required page numbers for submission
\pagestyle{plain}

\begin{abstract}
Crowd management is crucial for a smart campus. Popular methods are camera-based. However, conventional camera-based approaches may leak users' personally identifiable features, jeopardizing user’s privacy, which limits its application. In this work, we investigate using affordable light detection and ranging (LiDAR) technology to perform outdoor crowd management leveraging edge computing. Specifically, we aim to count the number of people on a walkway of a university campus. Besides privacy protection, LiDAR sensors are superior to cameras since their performance will not be compromised when the campus is not well-illuminated. We deploy LiDAR sensors on light poles to collect data from the crowd on the campus and leverage edge accelerators to process data locally. We proposed two different methodologies in this work: 1) a non-convolutional neural network (CNN)-based approach, using clustering and autoencoder, and 2) a CNN-based approach that first projects point clouds to 2D planes and then processes the projection with conventional CNNs. Our first approach relies on careful feature engineering, whereas our second approach does not require such effort. However, the CNN-based approach requires more computational power compared to our non-CNN-based approach. We evaluate both approaches comprehensively with our hand-labeled real-life data collected from campus. Our evaluation results show that the first method achieves an accuracy of 85.4\% whereas the second method achieves 95.8\%. Our CNN-based method outperforms existing solutions significantly. We also deploy our two models on an edge accelerator, TPU, to measure the speedup, leveraging this specialized accelerator. 
\end{abstract}


\section{Introduction}

% Also emphasize edge computing
% We do not send data back 
Outdoor crowd management is crucial for smart campuses.
\m{it is crucial for dumb campuses too. but why is it crucial? you did not really explain. what you really want to say is that crowd management is important and smart campuses can provide better solutions by utilizing IoTs and ML.}
Popular methods are camera-based, relying on video-processing algorithms that process footage captured by cameras. However, these camera-based solutions face performance degradation in low-light conditions and may leak users' private information.

To address the limitations of camera-based solutions, in this work, we investigate using light detection and ranging (LiDAR) sensors to perform outdoor crowd management. \m{do not come out as if you are the only one working on this.}
Specifically, we aim to detect and count the number of people on a university campus. Besides privacy protection, LiDAR sensors are better than cameras since their performance will not be compromised without light.\m{point cloud might also provide more info than 2D images?}

To solve this problem, we adopt a widely used two-stage pipeline that consists of clustering and classification. \m{why}
We propose two distinct classifiers to tackle the LiDAR-based crowd management problem: 1) a lightweight non-CNN-based approach utilizing an autoencoder and 2) a CNN-based approach. \m{why}
The non-CNN-based approach requires hand-crafted feature extraction whereas the CNN-based approach eliminates the need for hand-crafted feature extraction. However, it usually requires more training data to attain optimal results compared to our first classifier.

We deploy LiDAR sensors on strategically positioned smart BLUE light poles within our university campus to facilitate data collection from the surrounding crowd.
Our comprehensive evaluation employs our hand-labeled real-life data to determine the performance of both methods.
The evaluation results show that the first method achieves an accuracy of 85.4\% whereas the second method achieves 95.8\%. Furthermore, we deploy our models on two popular edge devices (Coral Dev Board and Nvidia Jetson Nano) to conduct on-device inference time measurement. To ensure the long-term stability of the deployment, we monitor the temperatures inside the smart poles and analyze the temperature fluctuations.

To ensure the cost-effectiveness for campus deployment, we use affordable LiDAR sensors for outdoor crowd management on the edge, presenting two unique challenges: 1) the need to process all the collected points in real-time on resource-constrained edge devices, and 2) these affordable LiDAR sensors yield a limited number of points compared to professional-grade sensors. The reduction in the number of points is particularly pronounced with increasing distance due to the diminishing surface area for light reflection. We need to carefully design the classifier and the feature extractor to ensure that they can capture the patterns within limited points.
\m{this paragraph should come out before you introduce your classifiers. then in your classifier discussion, you need to explain how they address these challenges.}

Our main contributions are summarized as follows: 1) To the best of our knowledge, we are the first to study crowd management using LiDAR-based methods on an actual university campus;\ming{what are the unique challenges of doing this on an actual campus} 2) Instead of using the existing open-source dataset, we collect and label all the data manually from a real-life university campus with affordable LiDAR sensors; \ming{what makes this dataset better? why can't we use existing data?} 3) We comprehensively evaluate the performance of the proposed methods.\ming{evaluation is not really a contribution}

The paper is organized as follows: Section~\ref{sec: background} introduces the background and related works; Section~\ref{sec: infrastructure} describes the infrastructure of the system; Section~\ref{sec: method} discusses the proposed methodology; Section~\ref{sec: experiment} presents the experimental results; Section~\ref{sec: deployment} discusses the deployment issues, and Section~\ref{sec: conclusions} concludes the paper.
\section{Background and Related Works}
\label{sec: background}

% \begin{itemize}
%     \item Crowd management background 
%         \begin{itemize}
%         \item Image-based and LiDAR-based crowd counting
%         \end{itemize}
%     \item Single class classification
%     \begin{itemize}
%         \item SVM
%         \item Autoencoder (AE)
%         \item 2D-CNN-based
%         \item 3D-CNN-based (point clouds classification) 
%     \end{itemize}
    
%     \item Clustering techniques
%     \begin{itemize}
%         \item DBSCAN
%         \item Centroid-based clustering
%         \item Limitations of centroid-based clustering in our task
%         \item Rationale of using DBSCAN for our work
%     \end{itemize}

% \end{itemize}

\subsection{Human Detection}
Outdoor crowd management leverages human detection. Distinguishing between humans and other inanimate provides extra information to understand the traffic flow in an outdoor area. It can help to make better decisions to design the road layout to better adapt to the flow. It can be helpful to enforce specific policies on campus, for example, walking-only zones in a specific campus area. There are two commonly used techniques: image-based and LiDAR-based. Image-based techniques are traditional and have great success. However, its performance degradation in low-light environments makes it less applicable in outdoor environments. Furthermore, images contain identifiable human features, which may incur privacy concerns. 

LiDAR sensors are an excellent alternative to solve the above limitations in traditional image-based methods. An important specification of LiDAR sensors is the ability to provide long-range and wide-angle laser scans. These laser scans produce point clouds that are usually very accurate and unaffected by lighting conditions.

However, human detection with LiDAR is still very challenging, especially when the person is far from the sensor. 
Since the point cloud becomes more sparse as the distance increases due to the decrease of the reflective surface area of a person. 
Recognizing humans from a relatively sparse set of points without additional color information is difficult. 
Our work focuses on using affordable LiDAR sensors in an outdoor environment, which is challenging since the number of points generated by our sensor is limited compared to professional-grade sensors, making it challenging to recognize humans from a distance. 

The traditional pipeline for human detection consists of several stages: clustering, feature extraction, and classification~\cite{yan2020online}.
The emerging techniques leverage deep neural networks to perform end-to-end learning. However, these methods are usually computationally intensive and require powerful GPUs. It is infeasible for real-time processing.

\subsection{Non-CNN-based Methods}
We formulate human detection as a single-class classification problem. 
%Because it depends on the distance between the LiDAR sensor and the object, the pattern of the human may vary. It is challenging to construct a training dataset that includes humans at all distances from the LiDAR sensor. 

\subsubsection{Autoencoder}
% To optimize our model for efficiency when deployed on an edge device, we consider a lightweight, non-CNN-based method. SVM has been used for human detection. However, the commonly used SVM implementation uses scikit-learn, which focuses on CPU-only platforms. To leverage the TPU in the Coral Dev Board, we formulate the human detection problem as a single class classification and opt for an autoencoder as our model. 

Autoencoder projects input vectors to a lower-dimensional vector and then reconstructs the data as the output. If the reconstruction error given a specific input is larger than a pre-defined threshold, this input is considered an anomaly. 
Autoencoder consists of an encoder and a decoder. The encoder consists of a number of neurons, forming a fully connected layer that converts the input vectors into hidden representations. The hidden representation’s dimension is determined by the hidden layer’s dimension. 

The resulting hidden representations are then sent to the output layer and converted to the original space, generating the reconstructed output. In the training phase, we optimize the autoencoder by minimizing the average reconstruction error, the error between the input vectors, and the reconstructed output. By minimizing an autoencoder’s construction error between the input vectors and the reconstructed output, we can obtain a set of optimal parameters for the autoencoder. 

In the test phase, as normal data in the test set matches the pattern the autoencoder learns during its training phase, it will produce a small reconstruction error, whereas the anomalous data will have a relatively higher reconstruction error. We can classify the anomalous data by defining a threshold for the reconstruction error: any test inputs with a reconstruction error lower than the threshold are considered to be normal data; otherwise, they are considered to be anomalous data.

\subsubsection{One-Class Support Vector Machine (OC-SVM)}

OC-SVM is a modified version of SVM for single-class classification. Unlike conventional SVM, which learns to maximize the separation of data points from two classes, OC-SVM considers the origin of the projected space as an outlier and finds a hyperplane that maximizes the margin between the origin and the other data points. 
%To separate the data points from the origin, the author proposes to solve the following problem~\cite{scholkopf2001estimating}:
% \yt{add the equation to explain. Doesn't seem to be that important here. }

\subsection{CNN-based Methods}

CNN-based methods can be categorized into two classes: projection-based methods and point-based methods. 

\subsubsection{Projection-based Methods}
Projection-based methods project the unstructured point clouds into 2D images. 

followed by their subsequent processing with Convolutional Neural Networks (CNNs) for classification. This process can be approached in two fundamental ways: first, by routing these images through multiple CNNs and subsequently merging the outputs of these CNNs, or second, by merging the images and then passing the resulting merged image through a single CNN. These methods first extract features from different views from the point cloud and then aggregate these features into a discriminative representation. \me{The flow is not good}

Multi-View Convolutional Neural Network (MVCNN)~\cite{su2015multi} \me{what are the more recent developments in this direction?} uses multiple 2D views of a 3D object for classification. It merges intermediate outputs from different CNNs into a single representation and then uses another CNN for classification. 

Hayton et al.~\cite{Hayton2020cnn} project point clouds onto 2D images using an occupancy grid. They stacked three projections to form an image, treating each projection as a channel in this image. This approach achieves an impressive accuracy of up to 98.8\% on a custom dataset collected on an unmanned aerial vehicle (UAV). 

Running multiple CNNs simultaneously places significant demand on resources, making it unsuitable for edge devices with limited computational resources. Inspired by Hayton et al.’s approach, we explore the use of a single CNN with multiple stacked projections for its efficiency and the potential for promising performance. 

\subsubsection{Point-based Methods}

Typical deep learning methods for 2D images cannot be directly applied to 3D point clouds due to their unstructured nature. As a pioneer work, PointNet~\cite{qi2017pointnet} directly takes point clouds as input and outputs class labels. Specifically, it leverages several MLP layers to learn features independently and extracts global features with a max-pooling layer. 
\section{Smart Blue Light Pole Infrastructure}\label{sec: infrastructure}

% \begin{itemize}
%     \item Entire on-campus deployment configuration
%     \item Lidar sensor setup and specs
%     \item Computing unit (Coral?)
%     \item Data collection
%     \item Other components
% \end{itemize}
In this section, we will delve into the infrastructure details. 

\ming{using past tense in this section is more appropriate}

\subsection{Configuration}

To achieve the optimal balance between cost and performance, we select the Ouster OS0 32-channel LiDAR sensor. Mounted two meters above ground on a smart BLUE light pole, this sensor continuously captures point cloud data from passersby, ensuring real-time data collection. The collected data is securely sent to the edge device within the light pole's protective compartment, facilitating efficient processing.
Table~\ref{tab: lidar_spec} and Table~\ref{tab: edge_spec} list the detailed specifications of the LiDAR sensor and the edge devices used. 

\begin{figure}[t]
    \begin{subfigure}{0.45\columnwidth}
        \centering
        \includegraphics[width=4cm]{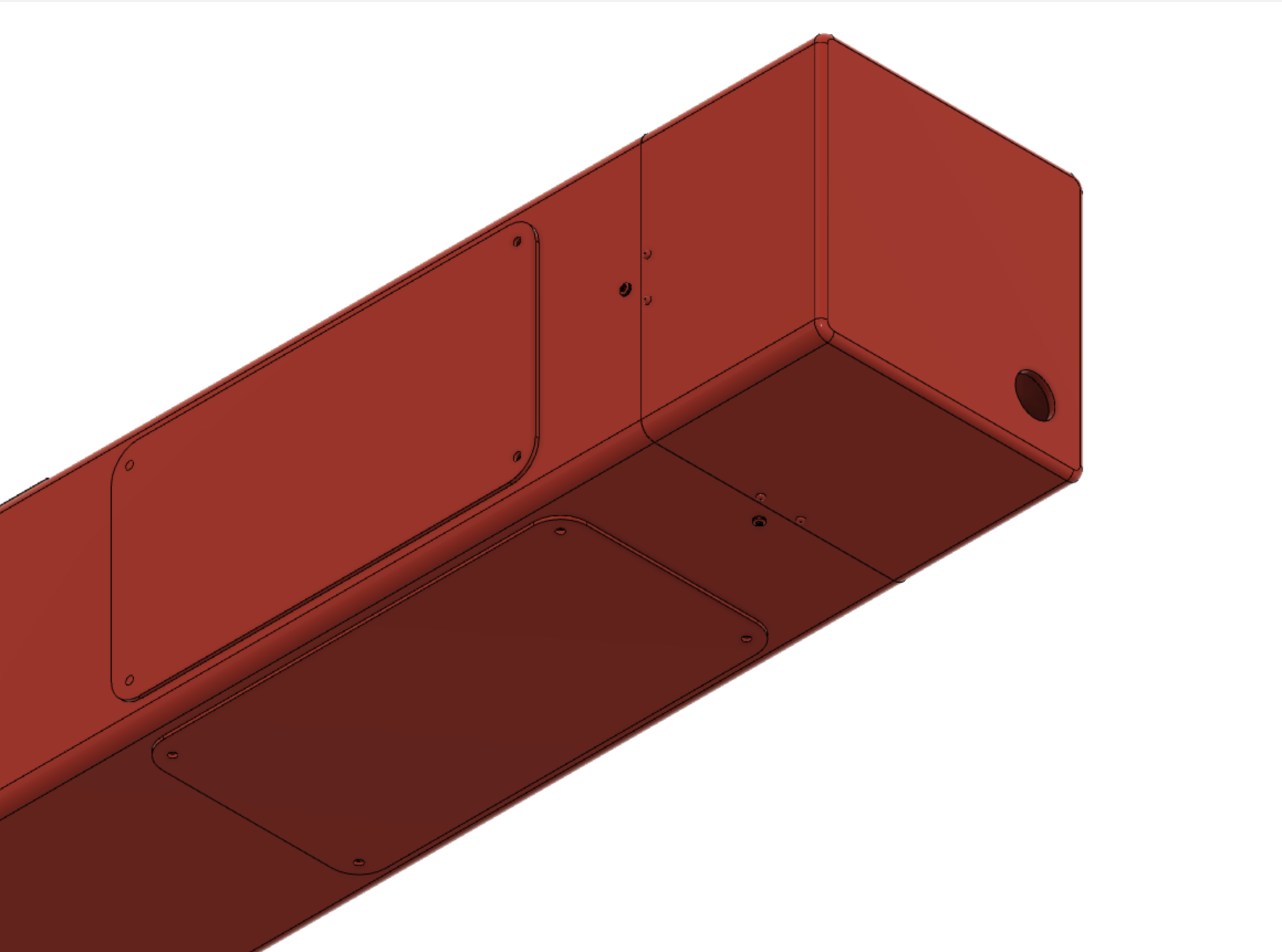}
        \caption{Pole diagram.}%\yt{Indicate where the compartment is.}
        \label{fig: pole}
    \end{subfigure}
    \begin{subfigure}{0.45\columnwidth}
        \centering
        \includegraphics[width=4cm]{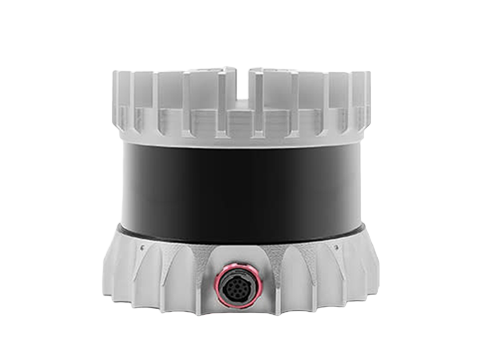}
        \caption{LiDAR sensor Ouster OS0.}
        \label{fig: lidar}
    \end{subfigure}
    \caption{Pole and LiDAR sensor.}
    \label{fig: pole_lidar}
\end{figure}

Fig.~\ref{fig: pole} shows the tilted view of a BLUE smart light pole without the LiDAR sensor. Inside the pole's compartment, the edge device is positioned for protection from heat and rain. Fig.~\ref{fig: lidar} is a front view of the LiDAR sensor mounted atop of the pole. 

\begin{table}[t]
\caption{Technical specifications for the LiDAR sensor\cite{lidarspec}.}
\label{tab: lidar_spec}
\centering
\begin{tabular}{@{}ll@{}}
\toprule
\multicolumn{2}{c}{Ouster OS0}                 \\ \midrule
Number of channels       & 32                      \\
Range                    & 35 m\\
Minimum range            & 0.3 m                   \\
Range resolution         & 0.3 cm                  \\
Horizontal resolution    & 512 (unit?)             \\
Horizontal Field of View & 360\degree              \\
Vertical Resolution      & 32 channels             \\
Vertical Field of View   & 90\degree                      \\
Rotation Rate            & 10 Hz                   \\
Points Per Second        & 655,360                 \\
Data Rate                & 66 Mbps                 \\ \bottomrule
\end{tabular}
\end{table}

\begin{table}[t]
\caption{Specifications for the edge devices.}
\label{tab: edge_spec}
\centering
\begin{tabular}{@{}lll@{}}
\toprule
               & Coral Dev Board                                                                      & Nvidia JetsonNano                                                   \\ \midrule
CPU            & \begin{tabular}[c]{@{}l@{}}Quad-core Arm \\ Cortex-A53\end{tabular}        & \begin{tabular}[c]{@{}l@{}}Quad-core Arm \\ Cortex-A57\end{tabular} \\
GPU            & \begin{tabular}[c]{@{}l@{}}Integrated GC7000 \\ Lite Graphics\end{tabular} & \begin{tabular}[c]{@{}l@{}}128-core Nvidia \\ Maxwell\end{tabular}  \\
AI accelerator & Google Edge TPU                                                            &                                                                     \\
Memory         & 1 GB                                                                       & 4 GB                                                                \\
OS             & Ubuntu 18.04                                                               & Mendel Linux                                                        \\ \bottomrule
\end{tabular}
\end{table}

\subsection{Data Collection}

We design our data collection methodology to simplify the region-of-interest (ROI) creation by focusing on objects within a 12-meter radius of a LiDAR sensor. Rather than collecting point cloud data from a full 360-degree LiDAR scan, we focus solely on approximately 90 degrees, capturing the walkways that connect popular areas on the campus and discarding the remaining data. \yt{The remaining 270 degrees of the data have no walkways, resulting in minimal student activity in those areas.}
Our fixed LiDAR sensors on smart blue light poles do not have\yt{the point cloud data synchronization issues encountered in previous studies when the LiDAR sensor is equipped on a moving device, such as an Unmanned aerial vehicle (UAV)~\cite{Hayton2020cnn}.}\ming{not sure you need to mention this}

We deploy LiDAR sensors in three key campus locations:\yt{gym, business school, and cafeteria. The gym is located near a major city street and has distinct traffic patterns, with peak attendance in the early morning and late afternoon or early evening, aligning with student's workout routines before or after school. 
The business school is located northeast of the gym, at an intersection of two streets, and observes high traffic during class hours.
The cafeteria is located to the north of the business school. The cafeteria traffic tends to peak during meal times with consistent attendance.} In this study,\yt{we focus on the LiDAR deployed at the cafeteria since the traffic is higher and more consistent than in the other two locations.} 

% Seems redundant here
% We utilize an Ouster OS0 32-channel LiDAR sensor, which has three modes (512, 1024, and 2048 horizontal resolutions at 10 and 20 Hz) and a fixed 32-channel vertical resolution. We collect all the data at 10 Hz using 512 horizontal resolution. The LiDAR sensor outputs 660K points per second. 

%%%%%%%%%%%%%%%%%%%% Citation index %%%%%%%%%%%%%%%%%%%%
%
%
% Title: CNN-based human detection using a 3D LiDAR onboard a UAV (hayton2020cnn)
% The motion of the UAV causes the raw point cloud data output by the LiDAR to not be synchronised. 
\section{Machine Learning-based People Counting}
\label{sec: method}

\subsection{Data Preprocessing}

%\ming{why do we need preprocessing? what are the challenges?}

Real-world LiDAR data can be noisy due to reflections from the ground or objects that are out of the sensor's effective range, which degrades the performance of the classifiers. Data preprocessing cleans the data to remove this noise and improve its quality. 
During the data preprocess step, we filter out the unnecessary data points reflected by the ground and define the Region of Interest (ROI). 

%Unlike prior studies considering varying ground levels~\cite{Hayton2020cnn}, our research specifically targets individuals walking along a flat walkway connecting different areas from a campus. \me{Thinking about removing this paragraph. Our setting eliminates the need to deal with the synchronization issues caused by the motion of a UAV. The ground level in our setting does not change, which is a simpler case. }

To mitigate the influence of ground-reflected noise, we employ a rule-based ground segmentation technique, keeping the points $p_i = (x_i, y_i, z_i) \in \mathbb{R}^3, i = 1, …, n$, where $z_i \ge z_{min}.$ ${z_i}$ denotes the coordinates of a point along the z-axis and $z_{min}$ denotes the minimum z value. The LiDAR sensor is fixed on the top of a three-meter-tall smart blue light pole, and its detection range along the z-axis extends from 0 to ${-3}$ meters. We empirically set $z_{min}$ to ${-2.6}$ meter.
\me{Removed. I meant different locations. I was thinking if the LiDAR was deployed at different location, then the ground-reflected noise may change. So we want to configure z value based on each deployment. However, we only have data collected from a single location. Maybe we don't need to discuss how to change the z value.} This approach assumes a flat ground level and a z-axis that is roughly perpendicular to the ground. It may result in the exclusion of small lower parts of objects.

% $z_{min}$ can be configured for each scenario. 

We empirically define our Region of Interest (ROI) based on specific criteria. The x-axis of the ROI restricts the distance to approximately 12 meters from the LiDAR sensors. This criteria ensures that individuals within this range are not affected by shadows cast by the blue light pole, enabling optimal collection of point clouds. Similarly, individuals too far from the LiDAR sensor result in weak reflective signals, limiting the number of points available for accurate classification. The y-axis covers the entire 5-meter wide walkway, allowing comprehensive coverage for data collection within the ROI.

\subsection{Clustering}
\label{subsec: clustering}

%\ming{why do we need clustering? what are the challenges?}

After data preprocessing, we apply clustering techniques to partition the resulting point clouds into distinct clusters. This serves two purposes: 1) noise reduction: even after preprocessing, the point cloud data may still contain noise. Clustering helps reduce this noise by grouping similar data points together; 2) handling multiple objects of interest. Clustering is particularly useful when dealing with scenes containing multiple objects of interest. It helps separate each object in the point cloud, making it easier to detect and classify multiple objects simultaneously.

Due to the unstructured nature of the point cloud data, the number of points within each cluster may vary. We employ different methods to ensure a fixed input size for the classifier. For the autoencoder and oc-svm approaches, we perform feature extraction on the resulting clusters to generate fixed-size feature vectors. In the subsequent section, we will discuss further details regarding the feature extraction step.
In the CNN2d approach, we enlarge the clusters to a predefined threshold and project them to create fixed-size 2D input images. Similarly, in PointNet, we enlarge the clusters to a predefined threshold without projection, directly passing them as input.
\me{Still working on the diagrams}\ming{we can discuss}

Following prior research\ming{what research}, we utilize the DBSCAN (Density-Based Spatial Clustering of Applications with Noise) algorithm, a widely used density-based clustering technique.\ming{to do what?}
The fundamental concept behind density-based clustering is to define a structure that accurately represents the underlying density of a given set of data points~\cite{kriegel2011density}.

Density-based clustering approaches differ from alternative methods, such as (Gaussian) mixture models, which assume a parametric distribution and typically assume clusters with convex shapes~\cite{jain1999data}. 
Other approaches, such as k-means clustering (k is a user-specified parameter, specifying the number of clusters to find\ming{not sure why you need to explain k here}), \yt{also contains assumptions about the cluster shape}~\cite{arbelaitz2013extensive}\ming{what does cluster shape mean? what assumptions?}.
In contrast, density-based clustering methods, including DBSCAN, do not rely on parametric distributions or variance measures.
\yt{Density-based clustering methods consider the high-density area of the data points as a cluster. This allows them to discover clusters with arbitrary shapes, including non-convex shapes, and handle noise as it treats the area of very low density of data points as outliers~\cite{hahsler2019dbscan}. Hence, density-based clustering methods offer flexibility and adaptability to different scenarios.} 

The DBSCAN algorithm works as follows:
\begin{itemize}
    \item Groups together data points close to each other based on a density criterion. It defines density as the number of data points within a specified distance ($\epsilon$) of a given point. A point is considered a core point if at least a minimum number of points ($minPts$) within its epsilon neighborhood. These core points serve as seeds for clusters.
    \item Uses reachability and connectivity to determine clusters. A point is considered directly reachable from another if it is within the epsilon neighborhood of that point. If a point is not directly reachable but can be reached by a series of steps within the epsilon neighborhood, it is considered reachable. This connectivity allows DBSCAN to find clusters of different shapes and sizes.
    \item Identifies noise points, data points that do not belong to any cluster. These points are usually isolated or lie in low-density regions. Border points are on the edge of a cluster and are reachable from a core point but do not have enough neighboring points to be considered core points themselves.
\end{itemize}

The DBSCAN algorithm requires two parameters: the neighborhood radius (\yt{$\epsilon$})\ming{use the greek letter?\yt{done}} and the density threshold (MinPts). Determining an appropriate value for MinPts often relies on the researchers’ domain knowledge and familiarity with the dataset, as there is no automatic method available. To find a suitable value for epsilon, we can plot the \mbox{k-nearest} neighbor distances (i.e., the distance of each point to its \mbox{k-th} nearest neighbor) in decreasing order. By analyzing this plot, we can identify an ``elbow'' or bend that indicates a transition from points within clusters (with smaller k-nearest neighbor distances) to isolated noise points (with larger \mbox{k-nearest} neighbor distances). The ``elbow'' in the plot indicates the optimal epsilon value of this input sample. In Section~\ref{sec: experiment}, we will examine the \mbox{k-nearest} neighbor distance plot for our dataset.

\subsection{Feature Extraction}\label{sec: feature_extraction}

Previous research has relied on distinct human shapes, such as leg patterns and head-to-shoulder profiles, to differentiate humans from other objects~\cite{arras2012range}. However, this approach faces challenges in extracting these partial features accurately at longer distances with reduced spatial resolution. To address this challenge, Kidono et al. proposed the slice feature approach~\cite{kidono2011pedestrian}. This approach transforms the point cloud data of a human into vertically connected slices or blocks, focusing on the 2D projections derived from these blocks. These 2D projections serve as features for distinguishing humans from other objects. While the slice feature approach has been proven effective in multiple studies~\cite{yan2020online, koide2019portable}, it is computationally intensive since it involves principal component analysis (PCA) along all three axes (x, y, and z) and eigenvector calculations during the 2D projection process.

Inspired by Kidono et al.'s slice feature approach, we propose a lightweight feature extraction method by eliminating the need for PCA and eigenvector calculations. Instead, we perform feature extraction on slices of each cluster of point cloud data along the z-axis. By dividing the point cloud into intervals of 0.02 meters (averaged human head length), we generate multiple slices for each cluster of point clouds. Subsequently, we extract features from each slice, capturing relevant information for human classification. This modified approach allows us to extract meaningful features from the point cloud data without the computational burden of PCA and eigenvector calculations.\ming{can you do an experimental comparison?} \me{Will try to do this.}

We employ the commonly used features for human classification following previous work~\cite{leigh2015person}. Considering the relatively small number of points output by our LiDAR sensor, we also calculate the basic statistics (such as mean, standard, deviation, etc.) of each feature to enhance the representation of each feature. The complete feature set derived from a cluster is a vector of 94 dimensions. 

\subsection{Classification Models}\label{classification_models}

In this section, we present our proposed models for human classification in LiDAR point clouds.  We propose two models: non-CNN-based with an autoencoder and CNN-based with CNN2d. \ming{summarize the complementary strengths of the two models}

\textbf{Autoencoder.} 
In the initial phase of our project, data scarcity, particularly with human classification, posed a significant challenge. \ming{present this as a method when data is scarce in general}
To address this, we treated the human classification task as a single-class classification problem and opted for an autoencoder as our model due to its lightweight and robust nature.

%We first formulate our human classification task as a single-class classification and investigate the commonly used autoencoders. 

We aim to deploy the model at the edge for real-time processing of incoming point cloud data. To achieve this, we use a relatively shallow, eight-layer, fully-connected layer only autoencoder consisting of a three-layer encoder, a bottleneck layer, a three-layer decoder, and an output layer. 
The bottleneck layer connects the encoder and decoder, determining the latent space dimension. To improve model performance and generalization, we apply dropout layers (ratio of 0.1) after each fully connected layer, excluding the bottleneck and output layers. 
To optimize the autoencoder's architecture, we utilize KeraTuner\ming{cite} for neural architecture search. We use it to determine the optimal number of neurons in the six layers, excluding the bottleneck and the output layers, by performing a grid search ranging from 16 to 128 neurons per layer. The resulting optimal autoencoder has a total of 26,384 parameters. Table~\ref{tab: ae_dim} shows the dimension of each layer in the autoencoder. 

\begin{table}[t]
\caption{Detailed autoencoder architecture.}
\centering
\begin{tabular}{@{}cll@{}}
\toprule
\multicolumn{1}{l}{Number} & Layer Type & Description \\ \midrule
1 & Fully-connected & Output size 104; dropout ratio 0.1 \\
2 & Activation & ReLu \\
3 & Fully-connected & Output size 72; dropout ratio 0.1 \\
4 & Activation & ReLu \\
5 & Fully-connected & Output size 124; dropout ratio 0.1 \\
6 & Activation & ReLu \\
7 & Fully-connected & Bottleneck layer size 8; dropout ratio 0.1 \\
8 & Activation & ReLu \\
9 & Fully-connected & Output size 76; dropout ratio 0.1 \\
10 & Activation & ReLu \\
11 & Fully-connected & Output size 84; dropout ratio 0.1 \\
12 & Activation & ReLu \\
13 & Fully-connected & Output size 76; dropout ratio 0.1 \\
14 & Activation & ReLu \\
15 & Fully-connected & Output size 94; dropout ratio 0.1 \\ \bottomrule
\end{tabular}
\label{tab: ae_dim}
\end{table}

\textbf{CNN2d.} The non-CNN approach mentioned earlier is lightweight and robust, but it relies on hand-crafted features, and finding the optimal set of features is challenging. To address this limitation, we consider CNNs as our next proposed model. CNNs can learn the features from raw data, and their feature extraction capability has shown remarkable success across diverse domains.

The output of the clustering algorithm discussed in Section~\ref{subsec: clustering} contains multiple clusters of points, and each cluster is an n x 3 array, where n is the number of points belonging to this cluster. To classify a cluster with the CNN-based method, there are two commonly used approaches:  1) use the cluster of points as input and directly pass them into a CNN model, such as PointNet; 2) convert the cluster of points to 2D ``images’’ and then pass the resulting 2D images to conventional CNNs for classification.

The first approach retains all the features, potentially achieving a higher accuracy than the second approach. However, it may require a large amount of training data for effective feature extraction due to point cloud data's sparse and unstructured nature. Furthermore, processing data in a 3D space requires much more computational power than in a 2D space. The constrained computation resources on the edge and the lack of training data due to the high labeling cost lead us to the second approach. 
\ming{provide a forward pointer}

To use the second approach, we need to convert point cloud data from a 3D space into a 2D space through projection. The commonly used projection method is the occupancy grid. However, this method is effective only with a large number of points and aims to produce low-resolution projection images. Note that if the number of points is limited, the resulting images may pose challenges in distinguishing valuable patterns from background noise.
Instead of projecting point clouds to a 2D plane using an occupancy grid, we directly project 3D point clouds by taking slices of the point clouds, minimizing the computational overhead incurred by the occupancy grid step. Slicing the xy plane shows the top-down view of the point cloud, slicing the yz plane shows the front view, and slicing the xz plane shows the side view.

% (Add visualization of different slices). 
%To evaluate the effectiveness of the projections, we conduct experiments to compare the performance of using these projections. 

The fixed-size input requirement of CNNs presents a challenge when dealing with clusters containing varying data points. To address this challenge, we need first to determine the appropriate number of points and then find a method to ensure that all the input samples conform to this number without compromising CNNs’ ability to learn the human pattern. 

Given the difficulty of identifying the significance of each point and the potential impact on CNN's ability to learn the human pattern, we choose to address the fixed input requirement challenge by adding more points to the point clouds. In order to determine the appropriate total number of points for the fixed-size input, we first analyze the maximum data points in our training dataset. Our findings show that the maximum number of points in our training dataset is 314. Since each point is represented as (x, y, z), the resulting point cloud data dimension is 314 x 3.

\begin{algorithm}[t]
\small
\caption{Enlarge dataset for CNNs. $maxRow$ is the maximum number of rows and $currentMaxRow$ of the entire dataset. $getMaxRow() $ retrieves $maxRow$ by iterating through the dataset.}
\label{algo: enlarge}
\begin{algorithmic}[1]
\Procedure{Enlarge}{}
    \State $maxRow \gets getMaxRow()$
    \State $objData \gets combineData()$
    \If {$isPerfectSqure(maxRow) \ne True$}
        \State $v \gets \sqrt{maxRow}$
        \State $v \gets ceil(v)$
        \State $maxRow \gets v^2$
    \EndIf
    
    \For {each file in the dataset}
        \State $d \gets read(file)$
        \State $numRows \gets CountRows(d)$
        \State $diff \gets maxRow - numRows$
        \State $newRows \gets sampleRows(diff, objData)$
        \State $d \gets concate(d, newRows)$
        \State $x, y, z \gets slice(d)$
        \State $img \gets stack(x, y, z)$
        \State $append(images, img)$
    \EndFor
    \State \textbf{return} $images$
\EndProcedure
\end{algorithmic}
\end{algorithm}

Since CNNs work on square images, we need to transform the point cloud into a square input image. To minimize the data points added while meeting the square image requirement, we set our target number of points to 324, which is only 10 points more than the maximum number of points in our training data. 
%This allows us to create an 18 by 18 image. 

To maintain a consistent count of 324 points within all our training samples, we introduce a controlled level of noise to each sample. Instead of sampling noise points from commonly used distributions (such as Gaussian), our approach samples these noise points from a distinct dataset, known as the ``ground dataset." This ground dataset is collected in environments where no humans are present. This sampling approach aims to minimize the impact of noise on CNN’s learning process. \m{ablation study? \yt{Still ongoing. Will add the results when they are ready}}

After increasing the number of points to meet our target, we create a 2D representation of the point cloud so that a conventional CNN can process them by projecting them to a 2D plane. This projection process generates three 2D representations of the point clouds: front-view, top-view, and side-view, resulting in three projections in the 2D space, each with dimensions of 324 x 2. To prepare these projections for input into our CNN, we utilize the reshape function to transform the projections from a size of 324 x 2 to 18 x 18 x 2. Following stacking the RGB channels in images, we stack all three 2D planes together to create an input image with a dimension of 18 x 18 x 6, which will be the input of our CNN2d model. \yt{Algorithm~\ref{algo: enlarge} summarizes the point cloud transformation algorithm.}

Our CNN model consists of three convolutional layers and two fully connected layers. Each convolutional layer is followed by a batch normalization layer and a ReLu activation layer. For all the convolutional layers, the size of the kernel filter is 3 x 3, and the stride is 1. 
Our proposed CNN2d consists of a total of 5 trainable layers and has a total of 62,114 parameters. Table~\ref{tab: conv2d} shows the detailed architecture of the CNN2d model.

\begin{table}[t]
\centering
\caption{Detailed CNN2d architecture. k denotes the kernel size, and s denotes the stride size.}
\begin{tabular}{@{}lll@{}}
\toprule
Number & Layer Type & Description \\ \midrule
1 & Convolution & Output size 32, k=3x3, s=1 \\
2 & Batch normalization & - \\
3 & Activation & ReLu \\
4 & Max Pooling & Kernel size 2x2 \\
5 & Convolution & Output size 64, k=3x3, s=1 \\
6 & Batch normalization & - \\
7 & Activation & ReLu \\
8 & Max Pooling & Kernel size 2x2 \\
9 & Convolution & Output size 64, k=3x3, s=1 \\
10 & Batch normalization & - \\
11 & Activation & ReLu \\
12 & Fully-connected & Output size 64 \\
13 & Fully-connected & Output size 2 \\
14 & Softmax & - \\ \bottomrule
\label{tab: conv2d}
\end{tabular}
\end{table}

% Each cluster goes through tailored steps based on different approaches.

% % To reduce the noise reflected by the ground with a rule-based filtering technique.  
% To make sure there are enough points to represent valuable patterns in the object of interest, we empirically define a region of interest. All the object present in this region generates a sufficient number of points to represent the pattern. 

%%%%%%%%%%%%%%%%%%%% reference index %%%%%%%%%%%%%%%%%%%%

% why is clustering used:
% 1) Reduce the number of points
% 2) Segment 
%
%

% Title: Online learning for 3D LiDAR-based human detection: experimental analysis of point cloud clustering and classification methods (yan2020online)
% "The traditional perception pipeline for object tracking consists of several stages, typically including segmentation (e.g. clustering), feature extraction, classification, data association, and position/velocity estimation."

% Title: A survey of deep learning-based object detection (jiao2019survey) (removed)

% "Point cloud based 3D object detection meets some challenges, the sparsity of LiDAR point clouds, highly variable point density, non-uniform sampling of the 3D space, effective range of the sensors, occlusion, and the relative pose variation."

% Title: Reflection Removal for Large-Scale 3D Point Clouds (yun2018reflection)

% Large-scale 3D point clouds (LS3DPCs) captured by terrestrial LiDAR scanners often exhibit reflection artifacts by glasses, which degrade the performance of related computer vision techniques.

%%%%%%%%%%%%%%%%%%%%%%%%%%%%%%%%%%%%%%%%%%%%%%%%%%%%%%%%%
\section{Experiments}\label{sec: experiment}
% \label{sec: exp}
% \begin{itemize}
%     \item Classification accuracy comparison
%     \item Impact of epsilon values on clustering performance
%     \item Accuracy of PointCloudCNN with varying model architectures
%     \item Robustness test?
%     \item Confusion matrix
%     \item Counting accuracy comparison
%     \item Inference time measurement on an edge device
%     \item Resource utilization measurement on an edge device
% \end{itemize}

\ming{use past tense}

We conduct various experiments to evaluate the performance of the proposed methods versus other baselines. The below section discusses the baseline models in this paper. 

\subsection{Baseline Models}

% In this section we describe the detailed model architectures of the two baselines. 

{\textbf{PointNet}. We consider PointNet~\cite{qi2017pointnet} since it is the first model to process point clouds and generate prediction labels directly. We adopt the PointNet implementation from previous work~\cite{qi2017pointnet} consisting of 64 layers and 747,947 parameters. Similar to our CNN2d approach, PointNet utilizes a fixed-size input and we provide such input by increasing the number of points in the point clouds to 324 x 3. PointNet consists of two essential components: the primary multiple-layer perceptron (MLP) network and the transformer net (T-net). The MLP network includes fully-connected layers with batch normalization and ReLu activation. The T-net consists of convolutional layers with batch normalization and ReLu activation. PointNet works well with a large number of points but demands significant computational resources. \ming{why?}
%The performance of PointNet in a small number of points from low-cost LiDAR is still unexplored. 
%We are also interested in understanding the inference time for PointNet on an edge device. 

% Table X shows the detailed model architectures of all three above models.
% (maybe just show the two proposed models. The PointNet is too big to show here)
\textbf{One-class SVM (OC-SVM)}. 
We consider OC-SVM because of its popularity in one-class classification. OC-SVM is a modified version of the conventional support vector machine (SVM) method, which has been popular for two-class nonlinear classification tasks. SVM transforms the nonlinearly separable data into a high dimensional space through a nonlinear map, or kernel, to achieve linearly separability of the data points within that space. To enable one-class classification, Bernhard Sch{\"o}lkopf et al.~\cite{scholkopf1999support} modified the conventional SVM, treating the origin of the high dimensional space as the only member of the second class.

\textbf{Implementation Details}. We implement the Autoencoder, CNN2d, and PointNet on TensorFlow 2.12 and conduct experiments on 4 Nvidia RTX 2080 Ti GPUs. We measure the inference time on a Coral Dev Board.\m{consider Jetson too} \yt{The CPU is NXP i.MX 8M SoC (consisting of a quad-core Cortex-A53 and a Cortex-M4F)}, the ML accelerator is a Google Edge TPU coprocessor, and the main memory size is 1 GB. For all the models, the total training epochs is 100 epochs. The optimizer is Adam and the learning rate is 0.001. The batch size is 32. 

\ming{explain your dataset and how it is used for training and testing}

% \subsection{Hyperparameters}

% For all the models, we train for 100 epochs to make sure the models are converged. The optimizer for the Autoencoder is Adam and the learning rate is 0.001. 

\subsection{Single Person Classification}

\begin{table}[t]
\centering
\caption{Single person classification comparison.}
\begin{tabular}{@{}lllll@{}}
\toprule
Model       & Accuracy & F1 Score & Precision & Recall \\ \midrule
Autoencoder & 0.8544   & 0.8534   & 0.7629    & 1.0000 \\
CNN2d       & 0.9587   & 0.9587   & 1.0000    & 0.9194 \\
CNN3d       & 0.9222   & 0.8991   & 1.0000    & 0.7403 \\
SVM         & 0.5600   & 0.3598   & 0.5620    & 1.0000 \\ \bottomrule
\label{tab:single_person_acc}
\end{tabular}
\end{table}

In the experiment, we evaluate the accuracy of our proposed models versus other baselines. 
Table~\ref{tab:single_person_acc} shows the results of the classification accuracy results. We evaluate commonly used metrics: accuracy, F1 score, precision, and recall. Accuracy calculates how many times a model correctly predicts if input point clouds belong to a human or an object. F1 Score is the harmonic mean of the precision and recall scores.  
On top of the F1 Score, we also show the Precision and Recall for a better understanding of the performance of our model. 
Among the four models, our proposed CNN2d method achieves the best accuracy and F1 scores. It has an accuracy of 95.87\% and the F1 score is 85\%. 
The autoencoder, albeit lightweight and easy to train, achieves only 85\% accuracy on the test dataset, which is 10\% worse than that of our CNN2d. This accuracy difference confirms CNN's capability of extracting useful features from 2D images formed by the projection of the point cloud data. PointNet’s accuracy is about 7\% better than the Autoencoder but 3\% worse than the CNN2d. This result confirms the feasibility of using PointNet on point cloud data produced by affordable LiDAR sensors. 
Even though PointNet directly processes the raw point cloud data, its accuracy is inferior to the proposed CNN2d method. One of the possible reasons could be that PointNet may require more training data than the proposed CNN2d method, since PointNet learns the pattern directly in 3D space. Given the limited hand-labeled training data in this work, PointNet fails to outperform our proposed 2D method. Finally, the OC-SVM baseline performs the worst, having only 56\% accuracy. 

\subsection{Optimal Clustering Parameters}

\ming{remind readers what the clustering is for}
We examine the \mbox{k-nearest} neighbor distance plot for our dataset in this section to determine the optimal Epsilon value. 
As mentioned in Sec~\ref{subsec: clustering}, we identify the ``elbow'' or bend that indicates a transition from points within clusters. Fig.~\ref{fig: single_elbow} shows the \mbox{k-nearest} neighbor distance plot for one of the training samples. In order to illustrate the “elbow” in the plot, we manually insert a horizontal line. \ming{no need to show this line} For this particular training sample, the “elbow” in the plot suggests that the optimal Epsilon value is 0.69. 

To determine the optimal Epsilon value for classifying all the samples, we analyze the neighbor distance plot generated from our training data, and identify the elbow values in the plot. \ming{confusing. single plot or many?} Fig.~\ref{fig: elbow_dist} shows the histogram of these elbow values, with a bin size of 0.02. 

From the histogram, we observe that the elbow value spans a wide range, varying from 0.04 to 0.18. Notably, the elbow value of 0.08 occurs the most frequently among all the values. Due to this wide range of elbow values, it is challenging to use a fixed value for achieving optimal performance across all samples. Additionally, the range of elbow values during deployment may exceed that of the training dataset. To address this challenge, we decide to compute the optimal epsilon value using the above-mentioned automated approach on each capture, ensuring optimal performance.
\ming{what automated approach?}
%However, computing elbow value for each capture incurs additional computation overhead compared to using a fixed value. We will discuss the trade-off in Section~\ref{}. 

\begin{figure}[t]
    \begin{subfigure}{0.45\columnwidth}
        \centering
        \includegraphics[width=4cm]{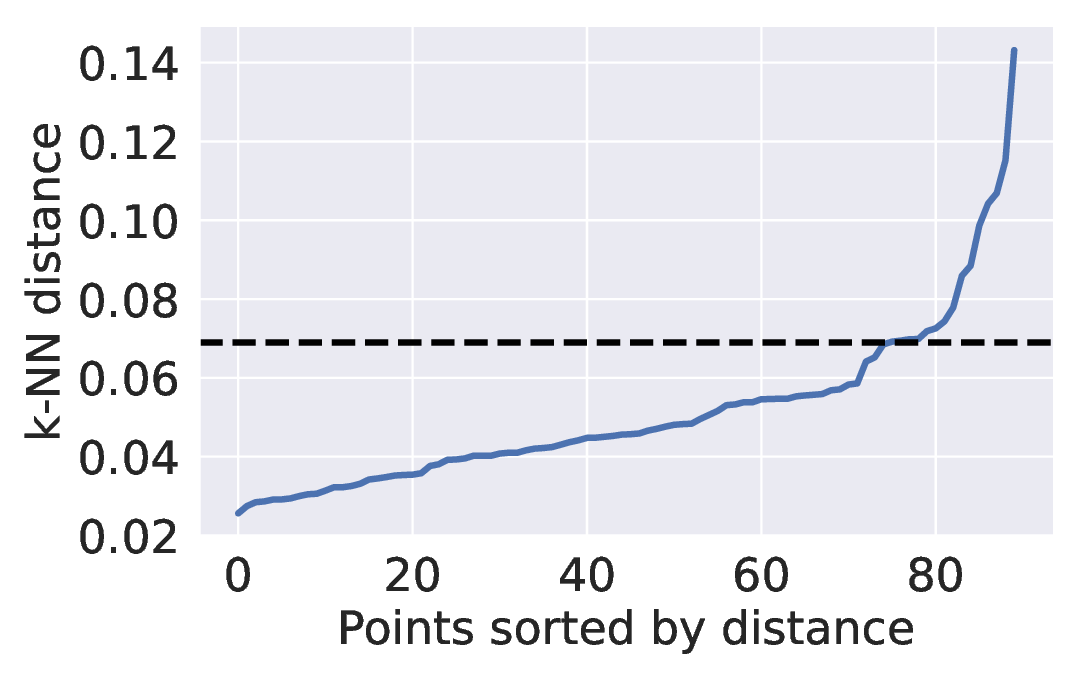}
        \caption{k-NN distance plot.}
        \label{fig: single_elbow}
    \end{subfigure}
    \begin{subfigure}{0.45\columnwidth}
        \centering
        \includegraphics[width=4cm]{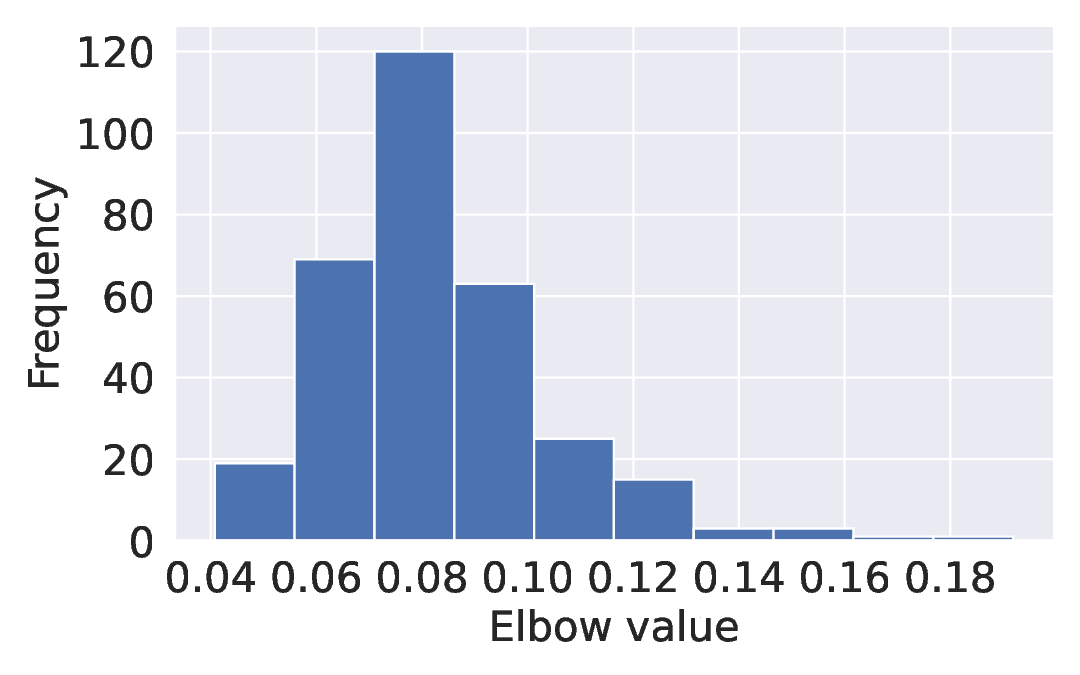}	
        \caption{Elbow values distribution.}
        \label{fig: elbow_dist}
    \end{subfigure}
    \caption{Elbow values.}
    \label{fig: elbow}
\end{figure}

% \subsection{Multiple people classification}
\subsection{Clustering Quality}

\begin{figure}[t]
    \begin{subfigure}{0.45\columnwidth}
        \centering
        \includegraphics[width=4cm]{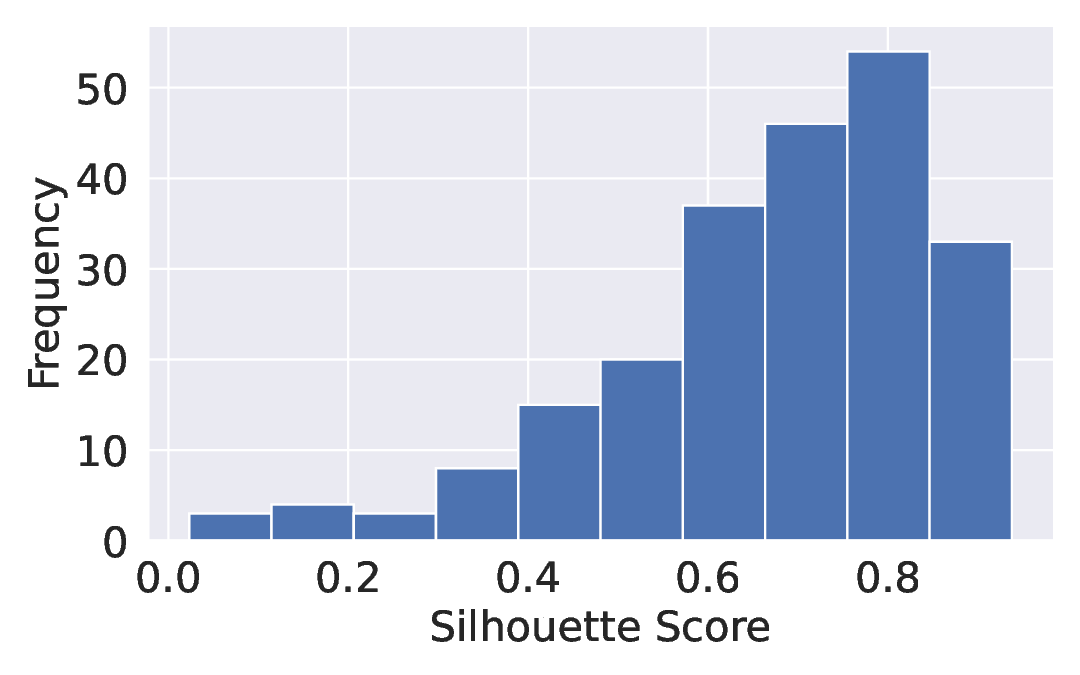}
        \caption{Silhouette scores.}
        \label{fig: Silhouette}
    \end{subfigure}
    \begin{subfigure}{0.45\columnwidth}
        \centering
        \includegraphics[width=4cm]{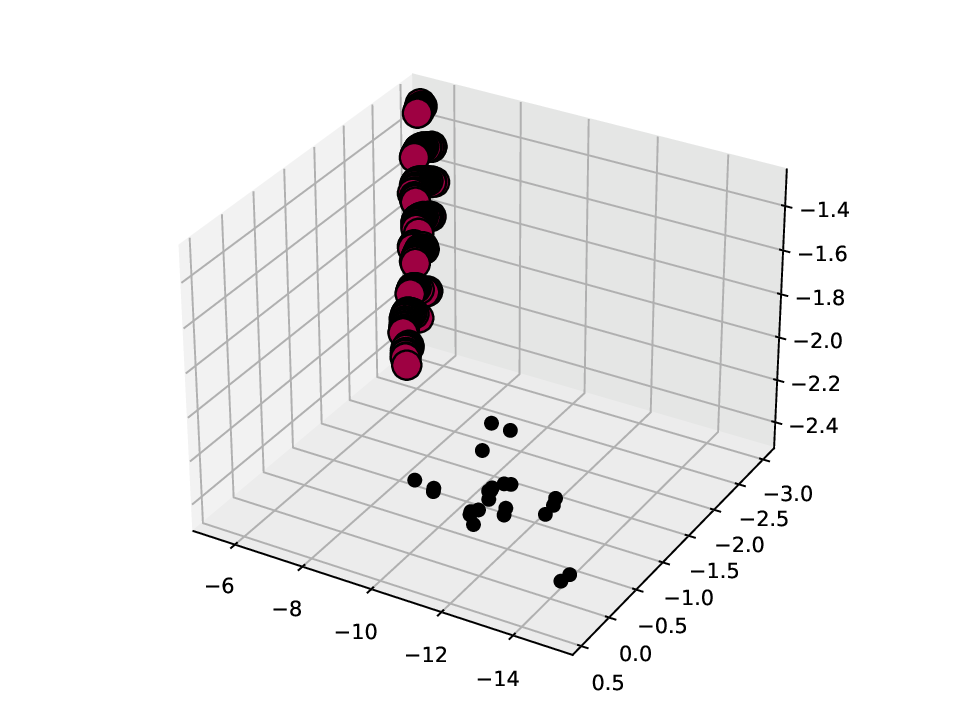}	
        \caption{Clustering visualization.}
        \label{fig: cluster_viz}
    \end{subfigure}
    \caption{Clustering quality.}\ming{fonts are too small}
    \label{fig: clustering_quality}
\end{figure}

In this section, we evaluate the clustering quality. To evaluate the clustering quality\ming{don't repeat the same words}, we use the Silhouette Coefficient~\cite{rousseeuw1987silhouettes}, which measures how well each data sample is clustered and it can be calculated using equation~(\ref{eq: sil}):
\begin{equation}
\label{eq: sil}
    S{(i)} = \frac{b(i) - a(i)}{ max\{a(i) , b(i)\}}, 
\end{equation}
where $a(i)$ denotes the mean intra-cluster distance for data sample $i$ and $b(i)$ represents the mean nearest-cluster distance for the same sample. The Silhouette Coefficient, ranging from -1 to 1, evaluates clustering quality, with -1 as the worst and 1 as the best~\cite{rousseeuw1987silhouettes}. 
% Note that Silhouette Coefficient is only defined if the number of clusters is larger than one. 

We compute the Silhouette Coefficient across our entire test dataset. Out of the 300 entries of test data, 77 entries form one or fewer clusters using DBSCAN. As the Silhouette Coefficient's definition requires a cluster count greater than one~\cite {rousseeuw1987silhouettes}, we exclude these entries and concentrate only on those with multiple clusters. Fig.~\ref{fig: Silhouette} shows the distribution for the Silhouette Coefficient scores. The Silhouette Coefficient has an average of 0.67 with a standard deviation of 0.18. This value confirms that our clustering parameters are reasonable for most of the input data using an adaptive epsilon value based on each input. To better understand the clustering quality, we also visualize the point clouds after performing DBSCAN in a 3D plot. Fig.~\ref{fig: cluster_viz} visualizes the clustering from one of the test data. The enlarged red circle markers correspond to the point clouds belonging to a person and the black circle markers represent the noise.

% To tackle the wide range of epsilon values, we calculate the optimal epsilon value based on each input during test time. 
% \begin{table}[t]
% \centering
% \caption{Multiple people accuracy comparison.}
% \begin{tabular}{@{}lllll@{}}
% \toprule
% Model       & Accuracy & F1 Score? & Precision? & Recall? \\ \midrule
% Autoencoder & 0.44827  &           &            &         \\
% CNN2d       & 0.266    &           &            &         \\
% CNN3d       & 0.53716  &           &            &         \\
% SVM         &          &           &            &         \\ \bottomrule
% \label{tab:multiple_people_acc}
% \end{tabular}
% \end{table}
% The results here do not make sense yet. 

\subsection{Edge Deployment}

\ming{this section does not belong in evaluation}

\ming{improve the flow of this section. connect your paragraphs.}

\ming{explain deployment on Jetson too?}

In this section, we discuss the details of deploying our models on the Coral Dev Board, an edge device installed in our smart blue light pole. By deploying our models on the Coral Dev Board, we take advantage of the powerful Tensor Process Unit (TPU) to accelerate on-device inference. 

% Table~\ref{} shows the detailed spec of the Coral Dev Board (add this in the infrastructure section). 

The Coral Dev Board utilizes a lightweight TensorFlow Lite Runtime interpreter, specifically designed for edge devices, to run TensorFlow Lite models in the tflite format. This interpreter supports two types of models: floating-point and quantized. floating-point models run on CPU whereas quantized models are optimized to run on TPU, maximizing inference speed.  

To convert a trained model into tflite format, we employ the Tensorflow Lite Converter. It is important to note that TensorFlow Lite is optimized for efficient inference, making it an ideal framework to run on resource-constrained devices like the Coral Dev Board. The TensorFlow Lite Converter can convert a trained model into both 32-bit floating-point and unsigned 8-bit integer format. During conversion, the converter freezes the model, fixing all the model weights and make it untrainable. This crucial step reduces the model size as it eliminates training-related components that are unnecessary for inference. 

To leverage the Coral Dev Board's TPU, which supports an 8-bit integer model only, we require model quantization. This process converts a 32-bit floating-point model into an 8-bit integer representation. To enable model quantization, we need to add an additional input argument to the TensorFlow Lite Converter, specifying the input and output type as unsigned 8-bit integer. 

The conversion from floating-point data to integers relies on the TensorFlow Lite Converter's capability to calibrate the quantization range. To determine this range, the TensorFlow Lite Converter requires users to provide a representative dataset. In our case, we randomly select 100 training images from our training data to serve as the representative dataset for calibrating the quantization range. With this representative dataset, the TensorFlow Lite Converter calculates the optimal quantization range for the model weights. Subsequently, the model converts into an unsigned 8-bit integer format while preserving its accuracy.

To take full advantage of the edge TPU, we utilize the Edge TPU Compiler on the unsigned 8-bit integer model. This compiler performs the crucial final optimization step on the quantized model, implementing specialized techniques tailored for the edge TPU, such as parameter data caching, to maximize the inference speed. 

The Edge TPU leverages an on-chip scratchpad memory (8 MB) for fast inference. 
Parameter data caching takes advantage of such memory to accelerate inference. During this final optimization step, the compiler assigns a unique ``caching token'' (a 64-bit number) to each model. When performing inference on a model, the Edge TPU runtime compares the caching token associated with the current data in the cache to the token of the model being run. If the tokens match, the runtime uses the cached data, avoiding the need for external memory access; if the tokens do not match, the runtime clears the cache and writes the new model's data into the cache for subsequent inference. This optimization step leverages the scratchpad memory to accelerate inference on the Edge TPU. 

\subsection{Inference Time on the Edge}

Fig.~\ref{fig: inference_time} shows the inference time comparison on the Coral Dev Board. We measure the inference time of a single\yt{LiDAR capture} since data is coming in a streaming fashion in a real-life scenario.\yt{Each LiDAR capture has a varying number of points. As elaborated in Sec~\ref{sec: feature_extraction}, we transform the point cloud into different sizes to suit the requirements of different models. For example, the Autoencoder operates on a 94-dimensional vector input, the CNN2d uses an input size of 18x18x6, and the CNN3d \ming{which one is 3d?} uses an input size of 324x3.}
%\m{what is in a single input? how big? why not process input in batch? \yt{we could process input in batch. I just thought that in a real-life scenario, the input comes one at a time. }} 
% We implemented all our neural network-based models, including Autoencoder, CNN2d, and PointNet) using TensorFlow version 2.12.0 and the oc-svm using Scikit-learn version 0.24.2.  

In our comparison of inference times between floating-point (32-bit) and quantized (8-bit) models, we find that quantization significantly improves inference speed. The autoencoder model achieves the highest speedup of 3.7X, reducing the inference time from 4.75 ms to just 1.25 ms per input sample. The CNN2d \ming{or CNN2D? be consistent}model follows with a speedup of 1.6X, resulting in an inference time of 1.07 ms per sample. The PointNet model shows a speedup of 1.3X, reducing the inference time from 53.31 ms to 41.88 ms. We notice that as model complexity increases, the speedup from quantization diminishes. Overall, our experiments demonstrate the effectiveness of 8-bit integer quantization in improving inference speed, with the autoencoder model achieving the most significant improvement.

To achieve real-time processing, a model must complete inference within a time window of 16 ms (assuming 60 fps). Fortunately, both of the proposed models have inference times below 16 ms, making them suitable for processing input samples in real-time. In comparison, while PointNet can run on Coral Dev Board without memory issues, its long inference time makes it impractical for real-time processing.

\begin{figure}[t]
\begin{subfigure}{0.45\columnwidth}
    \centering
    \includegraphics[width=4cm]{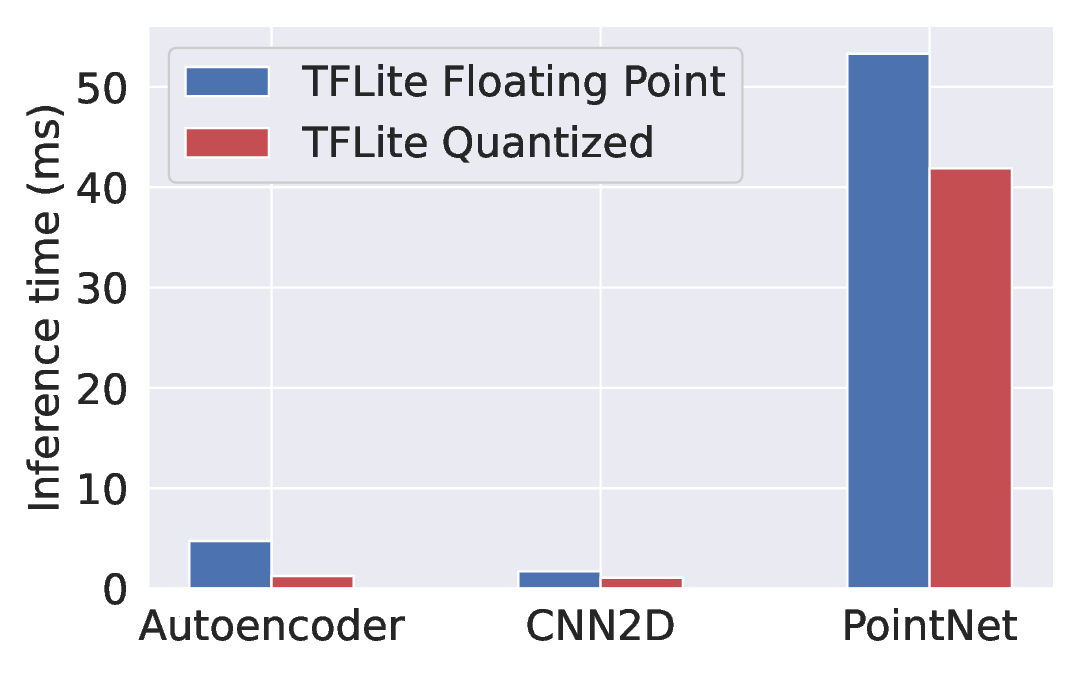}
    \caption{Inference speedup.}
    \label{fig: inference_time}
\end{subfigure}
\begin{subfigure}{0.45\columnwidth}
    \centering
    \includegraphics[width=4cm]{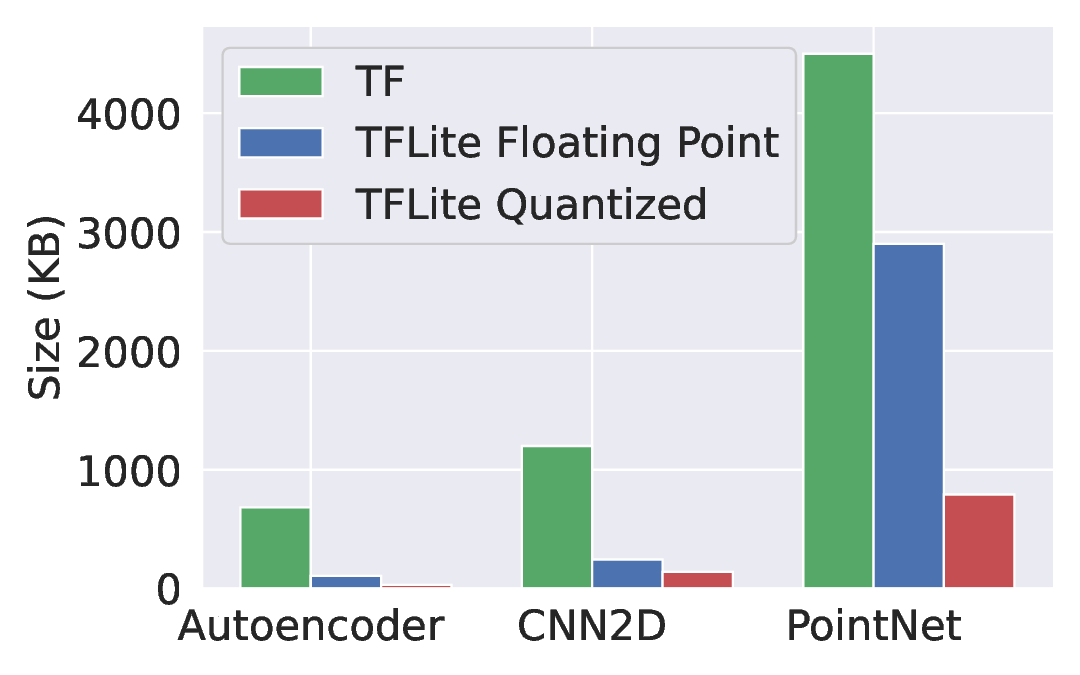}	
    \caption{Model size.}
    \label{fig: model_size}
\end{subfigure}
\vspace{-5pt}
\caption{Inference time and model size comparison on the Coral Dev Board. Fig. 2(a) demonstrates the inference times of three models. Fig. 2(b) compares the model size between the original model of TensorFlow (TF). The floating point model of TensorFlow Lite (TFLite floating point) and the quantized TensorFlow Lite model (TFLite Quantized) using 8-bit integers. The model size of the quantized Autoenoder is 31 KB, which is too small to be visible in Fig. 2(b). Note that the Coral Dev Board only supports tflite model so there is no TensorFlow model inference result in Fig. 2(a).}
\end{figure}

\subsection{Quantization on Model Size } \ming{upper case}

In addition to accelerating inference, quantization also significantly reduces the model size. However, similar to its impact on inference time, quantization’s effect on model size varies depending on the model architecture. Fig.~\ref{fig: model_size} illustrates the comparison of the model sizes for all three models. The autoencoder model\yt{reduces the size by 340\%,} from 107 KB to 31 KB. CNN2d\yt{reduces size by 170\%,} from 245 KB to 141 KB. \m{these models are much smaller than PointNet. How's the accuracy comparison?\yt{still working on getting the accuracy numbers. Will discuss that when the numbers are ready. }}Notably, PointNet has the largest size reduction\yt{of 360\%,} from 2.9 MB to 792 KB due to its complex architecture.\ming{contradicts to the next sentence} 
Surprisingly, the autoencoder model achieves a size reduction of 3.4X, which is only slightly lower than PointNet, indicating that quantization's impact is not solely dependent on model complexity.

\section{Deployment Issues}
\label{sec: deployment}

\begin{figure}[t]
    \centering
    \includegraphics[width=8cm]{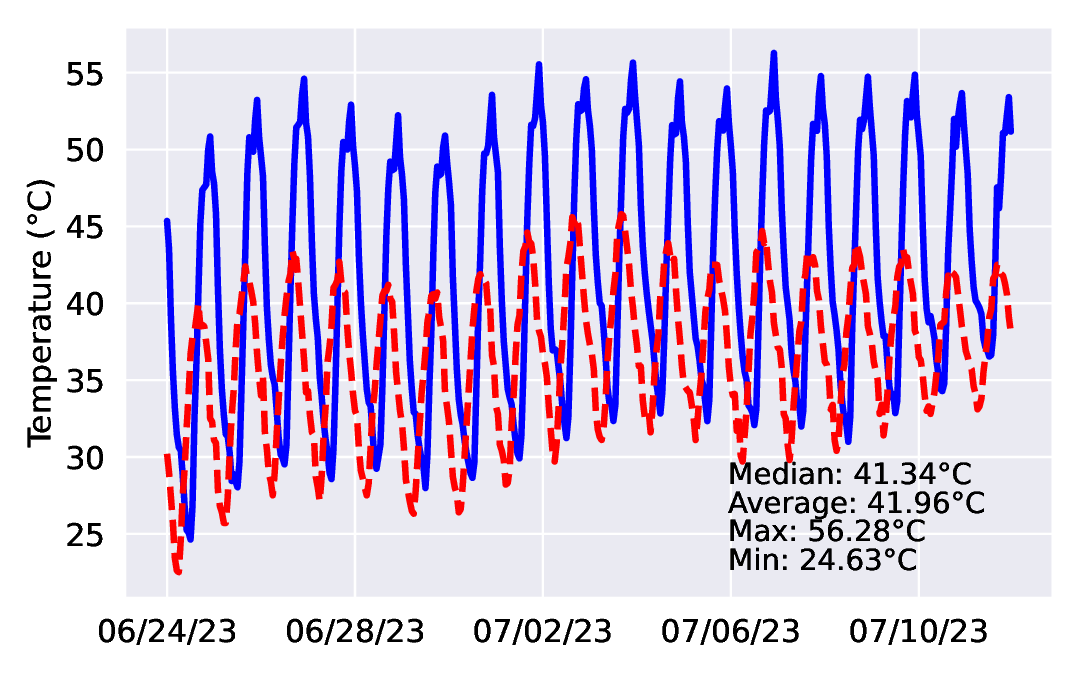}
    \caption{Pole temperature analysis.}\ming{you forgot to label the lines}
    \label{fig: pole_temp}
\end{figure}

We closely monitor the temperature of the pole to ensure it remains within the specified operational range of the Coral Dev Board, which is housed within the protective compartment of the pole. 
Given the large volume of temperature data collected during deployment of over a year, we focus on a specific period --- the scorching summer from June 24, 2023, to July 11, 2023. We collect temperature data daily, amounting to approximately 2500 data points per day. 

To gain a deeper insight into the impact of weather temperatures on the pole's temperature, we use weather temperature data obtained from Visual Crossing~\cite{visualcrossing}. The Visual Crossing's weather data provides hourly data, and to align the pole's temperature data with this format, we resample our pole temperature data at hourly intervals. 
% This synchronization ensures consistency between the two temperature sources. 

Figure~\ref{fig: pole_temp} illustrates the weather and the pole temperatures within this period. The pole temperatures exhibit variation each day closely aligned with weather temperature changes, with the maximum temperature reaching 57.81°C, the minimum at 21.00°C, and an average of 41.95°C. Notably, the highest recorded temperature was on July 6, 2023, whereas the lowest was observed on June 24, 2023. It is important to note that despite the slight elevation of the highest pole temperature above the recommended operational range of 0°C to +50°C~\cite{coralspec}, the Coral Dev Board continued to function without a single issue. Additionally, the difference between the pole temperature and the weather temperature remains consistent: during peak temperature, the pole temperature exceeds the weather temperature by approximately 10°C, while during the cooler periods, the difference is less than 5°C.

\ming{check with Krishna if there are any other worth-mentioning deployment issues}
\section{Conclusions}\label{sec: conclusions}

This paper investigates crowd management through LiDAR-based methodologies implemented on an actual university campus. 
Evaluation results indicate that the proposed non-CNN-based method and the CNN-based method achieve accuracies of 85.4\% and 95.8\%, respectively, on our hand-labeled real-life dataset. Our latency evaluations, utilizing a popular edge device (Coral Dev Board), demonstrate the real-time capabilities of both models. Furthermore, we conduct a detailed analysis of the weather temperature's impact on the edge device deployed within the pole. Our findings confirm the long-term stability of the device under diverse temperature conditions.

% \begin{itemize}
%     \item Accuracy comparison
%     \item Speed comparison
%     \item Effect on quantization
%     \item Deployment
% \end{itemize}

\bibliographystyle{ieeetr}
\bibliography{yitao}

\end{document}